\documentclass{article}

\def\makinhome{/home/makin}
\makeatletter
\begingroup\endlinechar=-1\relax
    \everyeof{\noexpand}%
    \edef\x{\endgroup\def\noexpand\homepath{%
        \@@input|"kpsewhich --var-value=HOME" }}\x
\makeatother

\ifx\homepath\makinhome
    \newcommand{\stydir}{../../stys}
    \newcommand{\bibsdir}{../../bibs}
    \newcommand{\tikzdir}{../../tikzpics/LRMwTLR}
    \newcommand{\figdir}{../../../jpgs/LRMwTLR}
\else
    \newcommand{\stydir}{stys}
    \newcommand{\bibsdir}{bibs}
    \newcommand{\tikzdir}{tikzpics/LRMwTLR}
    \newcommand{\figdir}{jpgs/LRMwTLR}
\fi

\providecommand{\stydir}{../stys}
\providecommand{\bibsdir}{../bibs}
\providecommand{\tikzdir}{../tikzpics}
\usepackage{%
	amsmath,amsfonts,
	amssymb,amsthm,graphicx,rotating,booktabs,
	versions, ifthen,siunitx,
}
\usepackage[table]{xcolor}
\usepackage{xcolor}
\usepackage{pgf,tikz,pgfplots} 
\usetikzlibrary{%
 	positioning,calc,patterns,shapes,arrows,intersections,backgrounds,fit,shadings,decorations.text,shadows,fadings,bayesnet,
}
\usepgfplotslibrary{colorbrewer,colormaps,fillbetween,groupplots}

\ifdefined\rvmacroize
\else
	\makeatletter
	\input{\stydir/jgmstd.sty}
	\makeatother
\fi
\newboolean{INCLFIGS} % default is false
\setboolean{INCLFIGS}{true}

\usepackage{caption,subfig}
\newcommand{\captioning}[2]{\caption{{\bf #1} {#2}}}
\ifthenelse{\boolean{INCLFIGS}}{
	\captionsetup[subfloat]{font={Large,sf},labelformat=simple,labelsep=none}
}{
	\captionsetup[subfloat]{labelformat=empty,labelsep=none}
}
\newcommand{\colorprovide}[2]{\@ifundefinedcolor{#1}{\colorlet{#1}{#2}}{}}
\usepackage[titletoc]{appendix}
\makeatletter
\input{\stydir/IMSL.sty}
\makeatother
\newcommand{\rEFH}{%
    \begin{figure}[!t]
    \centering
    \begin{tikzpicture}
        \tikzset{%
            latent/.append style={minimum size=1cm},
            obs/.append style={minimum size=1cm},
        }
        \node[obs] (u) {$\Genersuffstats{t-1}$};
        \node[below right=0.2cm and -1cm of u] (ulabel){$\generemission{patent=\genersuffstatsarg{t-1}} $};
        \node[latent, right=1.0cm of u] (x) {$\Generltnts{t}$};
        \node[above left=0.2cm and -1cm of x] (xlabel){$\generposterior{patent={\genersuffstatsarg{t-1},\generobsvsarg{t}}} $};
        \node[obs, below=1.4cm of x] (y) {$\Generobsvs{t}$};
        \node[left=0.0cm of y] (ylabel){$\generemission{} $};
        \edge[-] {u} {x};
        \edge[-] {x} {y};

        \plate{allvars}{(u)(x)(y)}{$T$};
    \end{tikzpicture}% this percent sign is necessary!!
    \captioning{The rEFH.}{All distributions factor completely.}%
    \label{fig:rEFH}%
    \end{figure}%
}

\newcommand{\rVAEbackward}{
    \begin{figure}[!t]  
    \centering
    \begin{tikzpicture}
        \tikzset{%
            latent/.append style={minimum size=1cm},
            obs/.append style={minimum size=1cm},
        }

        \node[obs] (u) {$\Genersuffstats{t-1}$};
        \node[below right=0.2cm and -1cm of u] (ulabel){$\generreversetransition{} $};
        \node[latent, right=1.0cm of u] (x) {$\Generltnts{t}$};
        \node[above=0.1cm of x] (xlabel){$\generprior{} $};
        \node[obs, below=1.4cm of x] (y) {$\Generobsvs{t}$};
        \node[left=0.0cm of y] (ylabel){$\generemission{} $};
        \edge[->] {x} {u};
        \edge[->] {x} {y};

        \plate{allvars}{(u)(x)(y)}{$T$};
        \node[below=0.0cm of allvars] (modellabel){generative model};
    \end{tikzpicture}% this percent sign is necessary!!
    \hspace{0.1in}
    \begin{tikzpicture}
        \tikzset{%
            latent/.append style={minimum size=1cm},
            obs/.append style={minimum size=1cm},
        }

        \node[obs] (u) {$\Datasuffstats{t-1}$};
        
        \node[latent, right=1.0cm of u] (x) {$\Recogltnts{t}$};
        \node[above left=0.2cm and -1cm of x] (xlabel){$\recogposterior{} $};
        \node[obs, below=1.4cm of x] (y) {$\Dataobsvs{t}$};
        
        \edge[->] {u} {x};
        \edge[->] {y} {x};

        \plate{allvars}{(u)(x)(y)}{$T$};
        \node[below=0.0cm of allvars] (modellabel){recognition model};
    \end{tikzpicture}% this percent sign is necessary!!
    \captioning{The rVAE.}{%
    Note that the generative and recognition models assert incompatible independencies.}
    \label{fig:rVAEbackward}%
    \end{figure}
}

\newcommand{\bouncingBallsGeneration}{
    \begin{figure}[ht!]
      \centering\label{tbl:bbresultsThirty}
      \floatsetup{justification=centering}
      \includegraphics[width=1\textwidth]{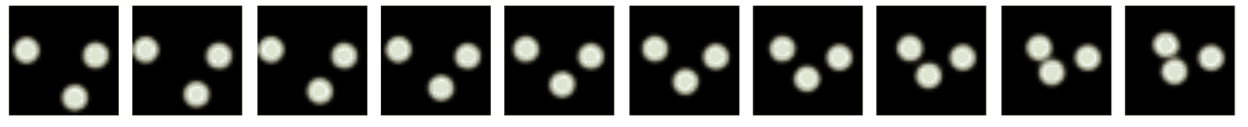} % Replace
      \includegraphics[width=1\textwidth]{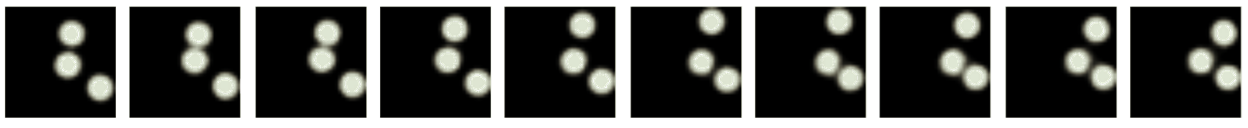} % Replace
      \caption{Snapshots of generated trajectory with the Bouncing Ball dataset}
      \label{fig:bouncingBalls}
    \end{figure}
}

\newcommand{\movingMNISTgeneration}{
    \begin{figure}[ht!]
      \centering
      \floatsetup{justification=centering}
      \includegraphics[width=0.8\textwidth]{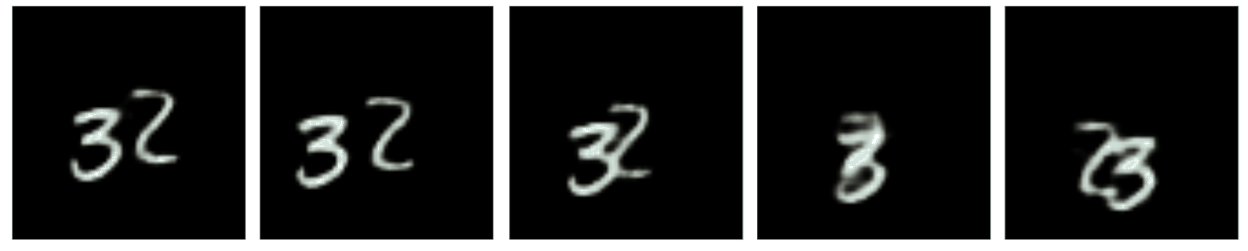} % Replace
      \includegraphics[width=0.8\textwidth]{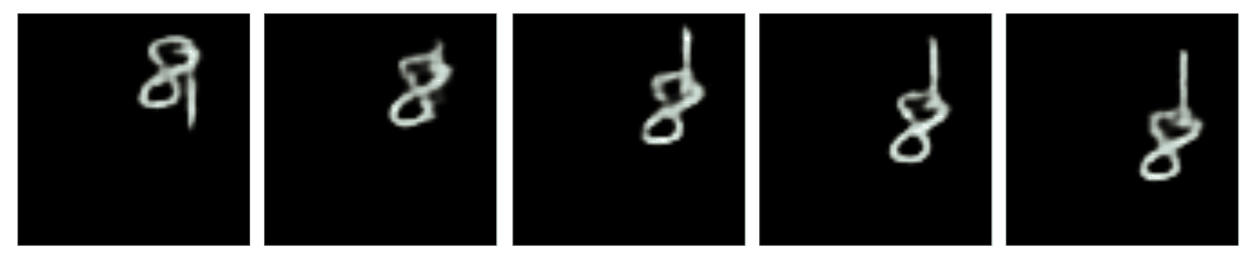} % Replace
      \caption{Snapshots of generated trajectory with the MovingMNIST dataset}
      \label{fig:MovingMNIST}
    \end{figure}
}

\renewcommand{\argtype}[1]{#1}
\usepackage{microtype}
\usepackage{booktabs} % for professional tables
\usepackage{hyperref}
\usepackage{graphicx}
\usepackage{floatrow}

\usepackage[accepted]{\stydir/icml2023}

\usepackage{mathtools}
\usepackage{xcolor}

\usepackage[capitalize,noabbrev]{cleveref}

\theoremstyle{plain}

\theoremstyle{definition}

\theoremstyle{remark}

\usepackage[textsize=tiny]{todonotes}

\def\suffstatsym{u}

\def\genersuffstatvar{\genermark{\suffstatsym}}

\rvmacroize[!][*]{dataobsv}\rvsequencemacroize{dataobsv}
\rvmacroize[!]{datasuffstat}

\rvmacroize[!]{genersuffstat}
\rvmacroize[!]{generltnt}\rvsequencemacroize{generltnt}
\rvmacroize[!]{generobsv}\rvsequencemacroize{generltnt}

\rvmacroize[!]{recogsuffstat}
\rvmacroize[!]{recogltnt}\rvsequencemacroize{recogltnt}
\rvmacroize[!]{recogobsv}\rvsequencemacroize{recogobsv}

\cmltmacroize[\generltntvar][*][\mu][\sigma]{trans}
\cmltmacroize[\generobsvvar][*][\mu][\sigma]{emiss}
\cmltmacroize[\recogltntvar][*][\nu][\upsilon]{recog}

\pgfkeys{distributions, adjust/.append style={/distributions/.cd, index=t}}
\pgfkeys{distributions, recog/.append style={/distributions/.cd, paramdisplay=;\altparams}}

\icmltitlerunning{Submission and Formatting Instructions for ICML 2023}

\begin{document}

\twocolumn[
\icmltitle{Learning Recurrent Models with Temporally Local Rules}



\icmlsetsymbol{equal}{*}

\begin{icmlauthorlist}
\icmlauthor{Azwar Abdulsalam}{yyy}
\icmlauthor{Joseph G.\ Makin}{yyy}
\end{icmlauthorlist}

\icmlaffiliation{yyy}{School of Electrical and Computer Engineering, Purdue University, West Lafayette, IN, U.S.A.}

\icmlcorrespondingauthor{J.G.\ Makin}{jgmakin@purdue.edu}
% \icmlcorrespondingauthor{Firstname2 Lastname2}{@www.uk}

\icmlkeywords{backpropation through time, RBM, VAE, local learning}

\vskip 0.3in
]



\printAffiliationsAndNotice{}  % leave blank if no need to mention equal contribution

\begin{abstract}
Fitting generative models to sequential data typically involves two recursive computations through time, one forward and one backward.
The latter could be a computation of the loss gradient (as in backpropagation through time), or an inference algorithm (as in the RTS/Kalman smoother).
The backward pass in particular is computationally expensive (since it is inherently serial and cannot exploit GPUs), and difficult to map onto biological processes.
Work-arounds have been proposed; here we explore a very different one:\ requiring the generative model to learn the joint distribution over current and previous states, rather than merely the transition probabilities.
We show on toy datasets that different architectures employing this principle can learn aspects of the data typically requiring the backward pass.
\end{abstract}

\section{Introduction}\label{sec:intro}
We consider the general context of trying to fit a model to sequential data, i.e.\ minimizing the KL divergence between the data distribution $\datamarginal{patent/\dataobsvsalltimearg} $ and the generative-model distribution $\genermarginal{patent/\generobsvsalltimearg} $.
The classic approach is to assume that the temporal correlations in the observed data $\Dataobsvs{t} $ can be explained by a latent state, $\Generltnts{t} $, that evolves according to a Markov chain.
% Before specifying a capacity for the latent state, the assumption is quite weak.
The assumption is not by itself very restrictive, and indeed this framework includes hidden Markov models (HMMs), linear-Gaussian dynamical systems (LGDSs), as well as modern variations like sequential VAEs.

Direct minimization of the marginal KL divergence, or equivalently cross entropy,
\begin{equation*}
    \ntrp{\datadistrvar\generdistrvar}{\Dataobsvsalltime}
        := 
    \def\integrand#1 {-\log\genermarginal#1 }
    \sampleaverage{patent/\Dataobsvsalltime}{\integrand}
\end{equation*}
generates a backward recursion:\ the gradient of this loss is coupled across time because the observations are.
When the dependencies are structured by a neural network, this recursion is called backpropagation through time (BPTT) \cite{Werbos1988}.
This is the approach taken, e.g., in various extensions of the restricted Boltzmann machine (RBM) to temporal data \cite{Sutskever2013,Boulanger-Lewandowski2012,Mittelman2014}.
For data arriving in real time, it is not clear how BPTT could be implemented biologically, since it would require a memory trace of the observed data.
And the algorithm is perforce serial, and therefore computationally expensive.

An alternative is provided by the expectation-maximization (EM) \cite{Dempster1977} algorithm, but it only avoids BPTT in certain special cases.
It is well known that the marginal cross-entropy is upper-bounded by the \emph{joint} cross entropy,
\begin{equation*}
    \begin{split}
        &\ntrp{\datadistrvar\generdistrvar\recogdistrvar}{\Recogltntsalltime,\Dataobsvsalltime} :=\\
        &\quad\def\integrand#1 {-\log\generjoint#1 }
        \sampleaverage{patent/\Dataobsvsalltime,latent/\Recogltntsalltime}{\integrand},    
    \end{split}
\end{equation*}
where the average is taken under a ``recognition model'' \cite{Dayan1995}, $\recogposterior{latent/\recogltntsalltimearg,patent/\dataobsvsalltimearg} $, as well as the data distribution.
The generative joint distribution ($\generdistrvar$) factors across time, due to the assumption that the dependencies in the observations ($\Dataobsvs{t}$) can be explained by the latent state ($\Generltnts{t}$).
This eliminates the need for BPTT in learning the \emph{generative model}.
In certain simple cases, like the HMM and LGDS, the generative model can be inverted in closed form with Bayes' rule, and used as the recognition model, in which case BPTT is not necessary at all.
However, the expectations under the recognition distribution ($\recogdistrvar$) still need to be computed with a forward and backward pass through the entire data sequence---famously, the forward-backward algorithm (HMMs) and Kalman filtering/RTS smoothing (LGDSs).
Similar considerations about computational efficiency and biological plausibility apply.
On the other hand, since the joint cross entropy is anyway a bound on the quantity we care about, the marginal cross entropy, the recognition distribution over latent states could simply be assumed (erroneously) to depend only on past observations---at the price of a looser bound.
This would amount to (e.g.) using the Kalman filter in place of the RTS smoother.

More critically, any sequential model much more expressive than the HMM or LGDS cannot be inverted with Bayes' rule.
Consequently, if the model is nevertheless to be trained with an EM-like framework---like sequential VAEs \cite{Saxena2021,Hafner2019}---the standard approach is to model the recognition distribution independently, with its own set of parameters, and learn them along with the generative model.
% although in diffusion models, the recognition distribution is parameter-free and not learned....
This again requires BPTT, this time through the recognition model (``encoder'' in the language of VAEs).

% This is the approach taken in some modern sequential models based on VAEs .
% It's clearly suboptimal, so they actually replace it with a multi-step predictor! and optimize it over various numbers of steps. But they still don't do a backward pass. But in any case they are essentially still--I think--grappling with the same problem.

Here we propose an alternative that still keeps learning temporally local.
In place of the standard generative model, a product over transition $\genertransition{} $ and emission probabilities $\generemission{} $, we propose to model at each time step the joint distribution of the current and previous states, along with the current observations.
More precisely, we model the distribution over \emph{the sufficient statistics for} the previous state, since these contain all the information about the preceding observation sequence (we make this precise below).
Intuitively, we require the model \emph{both} to yield good inferences about the hidden state, given the observations and previous state; \emph{and} to be a good generative model \emph{in reverse}.
We find that, together, these demands seem to enforce information flows in both directions, obviating the need for BPTT or a backward pass of inference.

To evaluate our approach, we concentrate on a toy dataset for which it is possible to reason about the dynamics, as well as two other simple datasets from the literature.

\section{Models}
\def\diagonalvar{d}
\def\generpast#1 {{%
    \assignkeys{distributions, gener, adjust, previous=\genersuffstatsarg{t-1}, #1}%
    \distribution{\previous\paramdisplay}
}}
\def\genertransition#1 {{%
    \assignkeys{distributions, gener, adjust, previous=\genersuffstatsarg{t-1}, #1}%
    \distribution{\latent|\previous\paramdisplay}
}}
\def\generreversetransition#1 {{%
    \assignkeys{distributions, gener, adjust, next=\genersuffstatsarg{t-1}, #1}%
    \distribution{\next|\latent\paramdisplay}
}}

\def\generjoint#1 {{%
    \assignkeys{distributions, gener, adjust, previous=\genersuffstatsarg{t-1}, #1}%
    \distribution{\latent,\previous,\patent\paramdisplay}
}}

\def\recogposterior#1 {{
    \assignkeys{distributions, recog, adjust, previous=\datasuffstatsarg{\index-1}, patent=\dataobsvsarg{\index}, latent=\recogltntsarg{\index}, parameters=\altparams, #1}%
    \distribution{\latent|\previous,\patent\paramdisplay}
}}
%%%%%%%%%%%%%
\cmltmacroize[\genersuffstatvar][*][\mu][\sigma]{trans}
\cmltmacroize[*][]{generic}

We focus on two architectures:\ a recurrent RBM \cite{Smolensky1986}, called the recurrent exponential-family harmonium (rEFH) \cite{Makin2015b}; and a recurrent VAE \cite{Rezende2014,Kingma2014}, rVAE.
For brevity we derive only the rVAE here, since the rEFH (\fig{rEFH}) has been derived elsewhere \cite{REFHDraft2015}.

\rEFH

\paragraph{Gaussian emissions.}
Consider the graphical model on the left in \fig{rVAEbackward}, parameterized by the distributions
\begin{equation}\label{eqn:generativeModel}
    \begin{split}
        \generprior{}
            &=
        \nrml{\vect{0}}{\mat{I}}\\
        \generreversetransition{}
            &=
        % \nrml{\TRANSITIONWTS\Generltnts{t}}{\cvrntranss}\\
        \nrml{\xpcttranss(\generltntsarg{t},\params)}{\vartrans\mat{I}}\\
        \generemission{}
            &=
        \nrml{\xpctemisss(\generltntsarg{t},\params)}{\varemiss\mat{I})}.
    \end{split}
\end{equation}
Our goal is to learn the parameters of this generative model for observed data $\Datasuffstats{t-1}, \Dataobsvs{t} \sim \datamarginal{patent={\datasuffstatsarg{t-1},\dataobsvsarg{t}}} $.
Classically, this would be carried out with EM, but for sufficiently complicated neural networks
$\xpctemisss(\generltntsarg{t},\params)$
and
$\xpcttranss(\generltntsarg{t},\params)$,
the final equation precludes computation of the posterior with Bayes rule.
Therefore to carry out an EM-like algorithm we make use of a \emph{recognition model} \cite{Neal1998} (see \fig{rVAEbackward}, right),
\begin{equation}\label{eqn:recognitionModel}
    \recogposterior{}
        =
    \nrml{
    \xpctrecogs(\recogsuffstatsarg{t-1},\dataobsvsarg{t})
        }{
    \cvrnrecogs(\recogsuffstatsarg{t-1},\dataobsvsarg{t})
    },
\end{equation}
that is likewise parameterized with neural networks (but functions of the observations rather than the latent variables), but that might not match the true posterior distribution under the generative model.
We further require the covariance to be diagonal:
\begin{equation*}
    \cvrnrecogs(\recogsuffstatsarg{t-1},\dataobsvsarg{t})
        :=
    \diag{\varrecogs(\recogsuffstatsarg{t-1},\dataobsvsarg{t})}.
\end{equation*}
From the generative and recognition models we can construct the free energy,
\begin{equation}\label{eqn:freeEnergyA}
    \begin{split}
        &\freeenergy(\params,\altparams)
            =
        \def\integranda#1 {\log\recogposterior#1 }
        \expectation{latent/\Recogltnts{t},previous/\Datasuffstats{t-1},patent/\Dataobsvs{t}}{\integranda}\\
        &\qquad-
        \def\integrandb#1 {\log\generjoint#1 }
        \expectation{latent/\Recogltnts{t},previous/\Datasuffstats{t-1},patent/\Dataobsvs{t}}{\integrandb},
    \end{split}
\end{equation}
which is an upper bound on the marginal cross entropy $\marginalXNTRP{patent={\Recogsuffstats{t-1},\Dataobsvs{t}}} $ \cite{Neal1998}, and has a tractable gradient.
(Note that the expectation is under the recognition distribution and the observed data.)
When \eqn{freeEnergyA} is minimized by gradient descent in the parameters $\params$ and $\altparams$ using the ``reparameterization trick'' \cite{Kingma2014} (that is, the pathwise gradient estimator \cite{Mohamed2020}), the resulting network is called a variational autoencoder.

\rVAEbackward

\eqn{freeEnergyA} can be rearranged to exploit the independence statements asserted by \fig{rVAEbackward}.
Substituting in the generative and recognition models from \eqns{generativeModel}{recognitionModel}, we find that
\begin{equation*}
    \begin{split}
        \freeenergy(\params,\altparams)
            &=
        % \def\integrand#1 {\log\recogposterior#1 - \log\generprior#1 - \log\generreversetransition#1 - \log\generemission#1 }
        % \expectation{latent/\Recogltnts{t},previous/\Datasuffstats{t-1},patent/\Dataobsvs{t}}{\integrand}\\
        %     &=
        \def\integranda#1 {%
            \assignkeys{distributions, gener, adjust, previous=\Genersuffstats{t-1}, #1}%
            % \sum_{\ncat}^{\Ncat}\log{\varrecog}_{\ncat}
            \sum_{\ncat}^{\Ncat}\log\stdrecog{(\ncat)}
        }
        \def\integrandb#1 {%
            \assignkeys{distributions, gener, adjust, previous=\Genersuffstats{t-1}, #1}%
            \xpctrecogs\tr\xpctrecogs + \stdrecogs\tr\stdrecogs
        }
        -\expectation{previous/\Datasuffstats{t-1},patent/\Dataobsvs{t}}{\integranda}
        +\frac{1}{2}\expectation{previous/\Datasuffstats{t-1},patent/\Dataobsvs{t}}{\integrandb}\\
            &\:\:\:\:\:+
        \def\integrand#1 {%
            \assignkeys{distributions, gener, adjust, previous=\Genersuffstats{t-1}, #1}%
            \Ncat\log\determinant{\vartrans}
            +
            \frac{\vectornorm{\previous - \xpcttranss}^2}{\vartrans}
        }
        \frac{1}{2}\expectation{latent/\Recogltnts{t},previous/\Datasuffstats{t-1},patent/\Dataobsvs{t}}{\integrand}\\
            &\:\:\:\:\:+
        \def\integrand#1 {%
            \assignkeys{distributions, gener, adjust, previous=\Genersuffstats{t-1}, #1}%
            \Obsvdim\log\determinant{\varemiss}
            +
            \frac{\vectornorm{\patent - \xpctemisss}^2}{\varemiss}
        }
        \frac{1}{2}\expectation{latent/\Recogltnts{t},previous/\Datasuffstats{t-1},patent/\Dataobsvs{t}}{\integrand}
        + c,\\
    \end{split}
\end{equation*}
with $c$ a constant term.
Since the emission cumulants, $\xpctemisss$ and $\cvrnemisss$, are complex (neural-network) functions of the latent variables, exact expectations under the recognition model $\recogposterior{} $ are intractable, and must be replaced with sample averages.
Likewise, since we have access to the data distribution only via samples, the expectations under $\Datasuffstats{t-1}$ and $\Dataobsvs{t}$ must also be replaced with sample averages.

\paragraph{Poisson emissions.}
We also consider the case where the observations $\Dataobsvs{t}$ are Poisson distributed and (still) conditionally independent:
\begin{equation*}
    \generemission{}
        =
    \prod_m^M \Pois{\lambda_m(\generltntsarg{t})},
    %     =
    % \prod_m^M \frac{\lambda_m(\Generltnts{t})^{\Dataobsv{t}{m}}\expop{-\lambda_m(\generltntsarg{t})}}{\dataobsvarg{t}{m}!}.
\end{equation*}
in which case the third term in the free energy becomes
\begin{equation*}
    \def\integrand#1 {%
        \assignkeys{distributions, gener, adjust, previous=\Genersuffstats{t-1}, #1}%
        \lambda_{\obsvdim}(\Generltnts{t}) - \Dataobsv{t}{\obsvdim}\log\lambda_{\obsvdim}(\Generltnts{t})
    }
    \sum_{\obsvdim}^{\Obsvdim} \expectation{latent/\Recogltnts{t},previous/\Datasuffstats{t-1},patent/\Dataobsvs{t}}{\integrand}.
\end{equation*}

\paragraph{Establishing recurrence.}
So far, despite the notation, the model is static.
We now identify the random variables.
In particular, we let $\Generltnts{t}$ be the latent state, $\Generobsvs{t}$ be the observations, and $\Genersuffstats{t}$ be \emph{the sufficient statistics for $\Generltnts{t}$}.
Recall that the sufficient statistics are any functions of the data---in this case, $\Recogsuffstats{t-1}$ and $\Generobsvs{t}$---that throw away no information about the underlying random variable---in this case, $\Generltnts{t}$.
Thus, if the recognition model matches the true posterior of the generative model, then the mean and variance functions 
$\xpctrecogs(\Recogsuffstats{t-1},\Dataobsvs{t})$
and 
$\cvrnrecogs(\Recogsuffstats{t-1},\Dataobsvs{t})$
are sufficient for $\Generltnts{t}$.
Accordingly, we define $\Recogsuffstats{t}$ to be the concatenation of these two functions.

Intuitively, the sufficient statistics provide a ``summary'' of 
$\Recogsuffstats{t-1}$ and $\Dataobsvs{t}$, or more precisely of their information about the latent state.
Since $\Recogsuffstats{t-1}$ is itself another summary, in this case of
$\Recogsuffstats{t-2}$ and $\Dataobsvs{t-1}$,
the argument can be extended recursively to claim that 
$\Recogsuffstats{t}$ summarizes \emph{all} of the preceding observations as they pertain to the latent state \cite{REFHDraft2015}, at least up to the capacity of this vector.
This makes it a good candidate to be explained by the generative model if we want it to learn how information propagates forward as well as backward in time.

\section{Experiments}\label{sec:methods}
We consider three experiments, one quantitative and two qualitative.

    \begin{figure}[ht]
        \raggedright%
        \providecommand{\figwidth}{0.98\textwidth}%
        \providecommand{\figheight}{1.5in}%
        \footnotesize%
        \provideboolean{CLEANXAXIS}\setboolean{CLEANXAXIS}{true}%
        \hfill% This file was created with tikzplotlib v0.10.1.
\begin{tikzpicture}
\provideboolean{CLEANTITLE}\ifthenelse{\boolean{CLEANTITLE}}{%
	\pgfplotsset{every axis post/.append style={title = {} }}%
}{}%
\provideboolean{NOLEGEND}%
\provideboolean{CLEANYAXIS}\ifthenelse{\boolean{CLEANYAXIS}}{%
	\pgfplotsset{every axis post/.append style={yticklabels = {} }}%
}{}%
\provideboolean{CLEANYAXIS}\ifthenelse{\boolean{CLEANYAXIS}}{%
	\pgfplotsset{every axis post/.append style={ylabel = {} }}%
}{}%
\providecommand{\figheight}{310pt}%
\pgfplotsset{compat=1.15}%
\providecommand{\thisXlabelopacity}{1.0}%
\provideboolean{CLEANXAXIS}\ifthenelse{\boolean{CLEANXAXIS}}{%
	\pgfplotsset{every axis post/.append style={xlabel = {} }}%
}{}%
\provideboolean{CLEANXAXIS}\ifthenelse{\boolean{CLEANXAXIS}}{%
	\pgfplotsset{every axis post/.append style={xticklabels = {} }}%
}{}%
\providecommand{\thisYlabelopacity}{1.0}%
\providecommand{\figwidth}{360pt}%

\definecolor{darkgray176}{RGB}{176,176,176}

\begin{axis}[
every axis x label/.append style={opacity=\thisXlabelopacity},
every axis y label/.append style={opacity=\thisYlabelopacity},
height=\figheight,
tick align=outside,
tick pos=left,
width=\figwidth,
x grid style={darkgray176},
xlabel={time (samples)},
xmin=-0.5, xmax=99.5,
xtick style={color=black},
y grid style={darkgray176},
ylabel={neuron ID},
ymin=-0.5, ymax=14.5,
ytick style={color=black}
]
\addplot graphics [includegraphics cmd=\pgfimage,xmin=-0.5, xmax=99.5, ymin=-0.5, ymax=14.5] {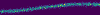};
\ifthenelse{\boolean{NOLEGEND}}{\legend{}}{}
\end{axis}

\end{tikzpicture}\hfill\par
        \setboolean{CLEANXAXIS}{false}%
        \vspace{-0.2in}%
        \renewcommand{\figheight}{1.5in}%
        \hfill% This file was created with tikzplotlib v0.10.1.
\begin{tikzpicture}
\provideboolean{CLEANTITLE}\ifthenelse{\boolean{CLEANTITLE}}{%
	\pgfplotsset{every axis post/.append style={title = {} }}%
}{}%
\provideboolean{NOLEGEND}%
\provideboolean{CLEANYAXIS}\ifthenelse{\boolean{CLEANYAXIS}}{%
	\pgfplotsset{every axis post/.append style={yticklabels = {} }}%
}{}%
\provideboolean{CLEANYAXIS}\ifthenelse{\boolean{CLEANYAXIS}}{%
	\pgfplotsset{every axis post/.append style={ylabel = {} }}%
}{}%
\providecommand{\figheight}{310pt}%
\pgfplotsset{compat=1.15}%
\providecommand{\thisXlabelopacity}{1.0}%
\provideboolean{CLEANXAXIS}\ifthenelse{\boolean{CLEANXAXIS}}{%
	\pgfplotsset{every axis post/.append style={xlabel = {} }}%
}{}%
\provideboolean{CLEANXAXIS}\ifthenelse{\boolean{CLEANXAXIS}}{%
	\pgfplotsset{every axis post/.append style={xticklabels = {} }}%
}{}%
\providecommand{\thisYlabelopacity}{1.0}%
\providecommand{\figwidth}{360pt}%

\definecolor{darkgray176}{RGB}{176,176,176}
\definecolor{darkorange25512714}{RGB}{255,127,14}
\definecolor{forestgreen4416044}{RGB}{44,160,44}
\definecolor{steelblue31119180}{RGB}{31,119,180}

\begin{axis}[
every axis x label/.append style={opacity=\thisXlabelopacity},
every axis y label/.append style={opacity=\thisYlabelopacity},
height=\figheight,
tick align=outside,
tick pos=left,
width=\figwidth,
x grid style={darkgray176},
xlabel={time (samples)},
xmin=0, xmax=99,
xtick style={color=black},
y grid style={darkgray176},
ylabel={angle (rad)},
ymin=-0.6507584782436, ymax=0.571198664289053,
ytick style={color=black}
]
\addplot [semithick, steelblue31119180]
table {%
0 -0.579886776686534
1 -0.58119635085675
2 -0.579569528752744
3 -0.577178533179291
4 -0.574152198604176
5 -0.570998746830776
6 -0.568031130232475
7 -0.562410667161667
8 -0.557757994464521
9 -0.551464042447797
10 -0.544625175316214
11 -0.537172952501768
12 -0.528382745619137
13 -0.519142441175962
14 -0.508101030591945
15 -0.496465094785299
16 -0.484273263602415
17 -0.471716903837636
18 -0.458926165378291
19 -0.445367648276047
20 -0.431297674090223
21 -0.416230470843328
22 -0.401919137368226
23 -0.386530656187995
24 -0.369769170811659
25 -0.353889686563009
26 -0.337402224151912
27 -0.320746680636076
28 -0.302341049656224
29 -0.283866007178355
30 -0.265191077661114
31 -0.246788762793053
32 -0.228259050747664
33 -0.208860956498023
34 -0.187579302369213
35 -0.167893108398697
36 -0.147638376297263
37 -0.128294638708358
38 -0.107777807506507
39 -0.0874110565751073
40 -0.0674631884536508
41 -0.0492233019229113
42 -0.030530248222191
43 -0.0131829186672709
44 0.00460233340853205
45 0.0232722338664853
46 0.0420046891660847
47 0.0596587836348515
48 0.0774976995108593
49 0.0957185040864605
50 0.113634580726055
51 0.131272104509058
52 0.149371917604591
53 0.168373214492752
54 0.186047733859043
55 0.203117120047541
56 0.220894881284696
57 0.238125695666779
58 0.255713031230625
59 0.271929275213194
60 0.286439720787833
61 0.301867593894574
62 0.316590770288837
63 0.32962947658806
64 0.343949842509657
65 0.357027536077634
66 0.370537510092712
67 0.382580359168491
68 0.395044760621766
69 0.406240776509688
70 0.41620686502887
71 0.425757316630789
72 0.435200454443237
73 0.443298890884137
74 0.451312040578068
75 0.456958782376961
76 0.46367845728319
77 0.46871651078108
78 0.474511505418245
79 0.47781724142726
80 0.481405479666737
81 0.483220488409318
82 0.484754798349888
83 0.484543188768656
84 0.484918574171868
85 0.48434148526899
86 0.483188881484227
87 0.481486199470943
88 0.478264391495037
89 0.474010667450896
90 0.468701801503326
91 0.464210919407291
92 0.458063759695984
93 0.452770160633389
94 0.445061873048899
95 0.436841457395846
96 0.427243279218303
97 0.418011247737112
98 0.408597595321721
99 0.398227818411582
};
\addplot [semithick, darkorange25512714]
table {%
0 -0.610865238198015
1 -0.6507584782436
2 -0.588081383430602
3 -0.540220958950626
4 -0.543998727894337
5 -0.590524934885299
6 -0.555655843492072
7 -0.544970154194148
8 -0.574777603288358
9 -0.548532050626789
10 -0.548532050626789
11 -0.531909867274462
12 -0.574777603288358
13 -0.548532050626789
14 -0.526590768601718
15 -0.520346609290235
16 -0.501598827043748
17 -0.491541707704526
18 -0.512913086300374
19 -0.478718880547016
20 -0.402049059834408
21 -0.392699081698724
22 -0.463758915529922
23 -0.299199300341885
24 -0.416277287432188
25 -0.326399236736602
26 -0.418879020478639
27 -0.311167272355561
28 -0.307999279763705
29 -0.241660973353061
30 -0.20943951023932
31 -0.345949191020305
32 -0.220837578823772
33 -0.175999588436403
34 -0.199466200227923
35 -0.16829960644231
36 -0.12965303014815
37 -0.131999691327302
38 -0.108044191790125
39 -0.0284951714611319
40 -0.0448798950512828
41 0.0373999125427356
42 -0.0448798950512828
43 0.0534284464896222
44 0.0224399475256414
45 -0.0130086652322561
46 0.0472419947908238
47 0.0854855143833959
48 0.115076653977648
49 0.112199737628207
50 0.105599753061842
51 0.14336633141382
52 0.190399554763018
53 0.135999681973584
54 0.129199697874905
55 0.213713785958489
56 0.203028096660565
57 0.264676304148591
58 0.23213738819629
59 0.237599444389144
60 0.293856455692923
61 0.367199141328677
62 0.288117844773667
63 0.279252680319093
64 0.369324136359514
65 0.394399077723394
66 0.339092540387469
67 0.418879020478639
68 0.421599014118111
69 0.461807615745083
70 0.415554583808174
71 0.48029361370671
72 0.439998971091007
73 0.448798950512827
74 0.498665500569808
75 0.488692190558412
76 0.46874557053562
77 0.49367884556411
78 0.490354408893645
79 0.478718880547016
80 0.482798871006223
81 0.436332312998582
82 0.571198664289053
83 0.482043317217481
84 0.512913086300374
85 0.491541707704525
86 0.448798950512828
87 0.478718880547016
88 0.463758915529922
89 0.553518705632488
90 0.498665500569808
91 0.436332312998582
92 0.436332312998582
93 0.503198823302261
94 0.460306615910592
95 0.43009899424146
96 0.460306615910592
97 0.482043317217481
98 0.441998966414148
99 0.43153745241618
};
\addplot [semithick, forestgreen4416044]
table {%
0 -0.602714409841598
1 -0.626550605795246
2 -0.611310248658963
3 -0.590529542714343
4 -0.581833631703653
5 -0.585906871665652
6 -0.574670488013929
7 -0.562459824041042
8 -0.563508778665552
9 -0.552762934556609
10 -0.545245154088376
11 -0.534805798026595
12 -0.544518654045717
13 -0.540492485916293
14 -0.529065866022976
15 -0.518819699747352
16 -0.505111470857581
17 -0.491918292724938
18 -0.492270154167113
19 -0.478733019446714
20 -0.444495597723134
21 -0.416480371064128
22 -0.425331784871226
23 -0.362962704457509
24 -0.372412375592809
25 -0.341801566054193
26 -0.357372941999124
27 -0.321208408890576
28 -0.301362634042298
29 -0.265360149860205
30 -0.226332419390975
31 -0.248577314016619
32 -0.222613568066746
33 -0.192558503715969
34 -0.180483637602573
35 -0.162975668631952
36 -0.142812796247315
37 -0.128764601980286
38 -0.112671072085905
39 -0.0737016517384701
40 -0.0528023798694299
41 -0.0185067284076139
42 -0.0157395074640192
43 0.0130728763198646
44 0.0254971748575097
45 0.0213347586644044
46 0.0369124302676731
47 0.0602298068986724
48 0.0833728523161226
49 0.100014651703705
50 0.109178691725375
51 0.12768898013152
52 0.156415366642308
53 0.160005225290559
54 0.154467931059996
55 0.180290832071349
56 0.192342646711041
57 0.220132843187664
58 0.231479389867787
59 0.241843943066737
60 0.270660121244131
61 0.314387190479833
62 0.310742171248169
63 0.31083940753541
64 0.343866293427407
65 0.370031204419243
66 0.367793111990346
67 0.39309909729302
68 0.410064251458331
69 0.435086073021217
70 0.434930308692548
71 0.455021203461273
72 0.454367505300421
73 0.453773812171165
74 0.472180300362796
75 0.481014909478923
76 0.47943723177448
77 0.484659822570044
78 0.488406134302415
79 0.483230573140472
80 0.482252619925766
81 0.47072228242124
82 0.501126834715341
83 0.494098229611206
84 0.50018480221399
85 0.496778219721881
86 0.472853545100611
87 0.471495646828246
88 0.46192759972713
89 0.493825124584983
90 0.494349030000855
91 0.474560409797903
92 0.459792938641776
93 0.472255527575671
94 0.461046277133236
95 0.442396421055663
96 0.443324293852975
97 0.4508160168069
98 0.441168602543716
99 0.430968433402634
};
\ifthenelse{\boolean{NOLEGEND}}{\legend{}}{}
\end{axis}%
\end{tikzpicture}\hfill\par
        \captioning{PPCs experiment (example).}{Upper panel: The observed data, consisting of 15 ``neurons'' responding to the position of an object moving with underdamped, linear, second-order dynamics.
        Lower panel: Position as a function of time:\ ground-truth (blue), as decoded from the observations (orange), and  as decoded from the reconstructed (denoised) observations (green).}
    \label{fig:LTIPPCs}
    \end{figure}

\paragraph{Probabilistic population codes (PPCs).}
We consider a simple data generator that starts with a one-dimensional, underdamped (oscillatory), second-order dynamical system that is driven by noise.
The position of this moving object is then ``reported'' by a crude model of a population of 15 neurons, with Gaussian-shaped tuning curves that uniformly tile the space of positions (interpreted to be angles).
The tuning curves provide the mean to a Poisson distribution from which spike counts are drawn \cite{Ma2006a,REFHDraft2015}.
This data set has three nice properties: (1) Since only position is observed, we expect models lacking a backward flow of information to learn only first-order dynamics, despite the fact that first-order systems cannot oscillate.
(2) The model is simple, but the relationship between observations and state is still nonlinear.
(3) Nevertheless, after applying the appropriate nonlinear transforms to the observations, a closed-form inference procedure is available in the form of a modified Kalman filter \cite{Beck2011,Makin2015b}.
This allows us to quantify how close to optimally position information is encoded in (or anyway can be decoded from) the latent state of our models.

\begin{table}[t]
\caption{Mean square errors (MSE) for recovery of position information on the PPC experiment.}
\vskip 0.15in
\begin{center}
\begin{small}
\begin{sc}
\begin{tabular}{lcccr}
\toprule
Model  &MSE   \\
\midrule
order 0       & \num{12e-4}\\
TVAE                          & \num{9.5e-4}\\
TRBM*                      & \num{6.0e-4}   \\    
KF-1          & \num{5.8e-4}\\
\textbf{rVAE}                  & \num{5.3e-4}\\
\textbf{rEFH}                & \num{3.3e-4}\\
RTRBM*                       & \num{3.1e-4}  \\
KF-2              & \num{2.2e-4}\\
\bottomrule
\end{tabular}
\end{sc}
\end{small}
\label{tbl:PPCresults}
\end{center}
\vskip -0.1in
\end{table}

\begin{table}[t]
\caption{Mean square errors (MSE) for the bouncing-ball dataset.
All cases use a single trajectory for each batch}
\vskip 0.15in
\begin{center}
\begin{small}
\begin{sc}
\begin{tabular}{lccr}
\toprule
Model & MSE \\
\midrule
order 0     & 0.0120\\
TRBM        & 0.0124\\
rEFH        & 0.0067\\
RTRBM       & 0.0059\\  
\bottomrule
\end{tabular}
\end{sc}
\end{small}
\label{tbl:BBresults}
\end{center}
\vskip -0.1in
\end{table}

\paragraph{Bouncing balls.}
Three balls move at constant speeds and (see \fig{bouncingBalls}) bounce off each other and the perimeter of the frame with complete energy conservation \cite{Sutskever2007}.
As with the preceding data set, we expect models that lack a backward flow of information to fail to learn that the velocities are constant (until collisions), since this requires learning a second-order dependency.

\paragraph{MovingMNIST.}
An extension of MNIST, it introduces dynamic elements with sequences of frames. Digits exhibit diverse motions like translations, rotations, and scaling. Interactions between multiple digits and collisions with the boundaries simulate real-world scenarios. It serves as a benchmark for video analysis, motion prediction, and object tracking. We expect similar results to the bouncing balls.

\paragraph{Details of the VAEs.}
For our experiments involving the bouncing balls and the MovingMNIST dataset, we employ an architecture inspired by the DCGAN \cite{Radford2016}.
In particular, our encoder network (recognition model) and decoder network (generative model) are composed, respectively, of convolutional and deconvolution layers.
However, only the images $\dataobsvs{t}$ pass through the convolutional layers of the encoder, after which they are concatenated with the (unprocessed) previous sufficient statistics, $\recogsuffstats{t-1}$, i.e.\ the vector of posterior means and variances (see \eqn{recognitionModel}) at time $t-1$.
This combined vector is then passed through a fully-connected layer with ReLU \cite{Nair2010} activations.
To obtain the cumulants of the posterior distribution at time $t$, the output is then passed through two different fully-connected layers, one for the mean and the other for the variance. 

For the PPC dataset, both the encoder and decoder are instead composed of two fully connected layers.
As in for the bouncing-ball and the MovingMNIST datasets, the PPC observations are passed through this portion of the encoder and then concatenated with the previous sufficient statistics.
This combined vector is subsequently passed through a fully connected layer with non-linear activation.

All updates to models were made by stochastic gradient descent with AdaM optimization \cite{Kingma2013}.

\section{Results}
To evaluate models trained on the PPC dataset, we compute the posterior mean under the recognition model at all time steps of a ``trajectory.''
From these we compute the expected value of the observations under a Poisson emission.
Critically, this ``updated'' version of the observations now contains information from the previous sufficient statistics (see again \fig{rVAEbackward}) and therefore can---if the model is good---provide a better estimate of the underlying position being encoded.
This estimate is computed with the ``center of mass'' of the population, i.e weighting each neuron's preferred angle by its number of spikes and normalizing, which is the optimal estimate of the encoded variable \cite{DayanAbbott}.
\fig{LTIPPCs} shows a typical example.

To quantify performance, we compare against the optimal inference algorithm, the Kalman filter (see \sctn{methods}), when applied to state-space models acquired wih EM.
In particular, we acquire both a first- and a second-order model.
\tbl{PPCresults} shows that, as expected, models trained according to our procedure (\textbf{rVAE} and \textbf{rEFH}) come close to the optimal learned Kalman filter (KF-2).
Removing the distribution over $\Genersuffstats{t-1}$ turns the rEFH into a model introduced by \cite{Sutskever2007} as the TRBM.
Accordingly, we call the corresponding variant of the VAE the TVAE.
Although these models are better than decoding from the uncorrected observations alone (``order-0''), they cannot outperform the optimal first-order model (KF-1).
Allowing BPTT in the TRBM turns it into the RTRBM \cite{Sutskever2009,Sutskever2013}, which restores ability to learn second-order dynamics, as expected.

In the case of bouncing balls, we ask how well each model can predict the next frame.
To make predictions, we run the model forward on an input sequence up till time $t$, and then use clamped Gibbs sampling to establish the next hidden state, $\Generltnts{t+1}$.
Finally, we noiselessy generate a ``sample'' $\generobsvs{t+1}$, and compare it to the actual next frame, $\dataobsvs{t+1}$.
Results are reported in \tbl{BBresults}.
Again the rEFH performs nearly as well as the RTRBM, despite omitting BPTT; whereas predictions from the TRBM, trained without BPTT or a distribution over the previous hidden state, are much worse, close to the prediction provided by the previous frame (``order 0'').
This is consistent with the PPC results:
Despite moving at a fixed, non-zero speed, the balls \emph{ average} zero velocity over long trajectories.
Consequently, the optimal first-order prediction is the same as the optimal zeroth-order prediction:\ the previous frame.
So it again appears that the TRBM learns first-order dynamics while the rEFH and RTRBM learn something second-order.

In our experiments to date, clamped Gibbs sampling under the rVAE does not converge, so we were unable to generate predictions from this model.
However, rVAE (like the rEFH) does not require Gibbs sampling for generation \emph{backwards in time}, so we instead evaluate it qualitatively by examining generated trajectories.
\fig{bouncingBalls} shows frames from one such trajectory.
The three balls follow anticipated trajectories before and after colliding with each other and the wall, demonstrating that the model has captured both the constant movement directions in the absence of collisions and the effects of those collisions, which suggests that it has learned something second-order.

\bouncingBallsGeneration

For MovingMNIST dataset, trajectories generated by the rVAE correctly bounce of walls (\fig{MovingMNIST}) and pass through one another.

\movingMNISTgeneration

\section{Conclusions}
The proposed approach of learning the joint distribution over current and previous states presents a promising alternative to computationally expensive backpropagation through time.
This study successfully verified the effectiveness of this principle on a toy dataset, demonstrating its ability to capture essential aspects that otherwise require BPTT.
Furthermore, qualitative evaluations with more sophisticated datasets---the bouncing-ball dataset and MovingMNIST---highlight the models' capacity to effectively capture non-linear second-order dynamics.
These findings showcase the potential of the proposed method to overcome computational challenges and open up new possibilities for more biologically inspired modeling of sequential data.

Nevertheless, although intuitively plausible, our procedure lacks a rigorous mathematical justification.
It also remains to scale the procedure up to more challenging datasets.

\bibliographystyle{icml2023}
\bibliography{%
    \bibsdir/machinelearning,
    \bibsdir/math,
    \bibsdir/nonpapers,
    \bibsdir/misctech,
    \bibsdir/compneuro
}

\end{document}